\documentclass{article}
\PassOptionsToPackage{table}{xcolor}
\usepackage{iclr2027_conference}
\iclrfinalcopy
\usepackage{newtxtext,newtxmath}
\usepackage{booktabs}
\usepackage{amsmath}
\usepackage{array}
\usepackage{tabularx}
\usepackage{float}
\usepackage{graphicx}
\usepackage[hidelinks]{hyperref}
\usepackage{microtype}
\usepackage{url}
\newcolumntype{Y}{>{\raggedright\arraybackslash}X}
\definecolor{auditblue}{RGB}{226,239,248}
\title{CIViC: Evidence Gates for Audio-Conditioned\\Visual Prediction on Proxy Tasks}
\author{
\begin{tabular}{c}
Mengzhe Geng$^{1}$\thanks{Corresponding authors: Mengzhe Geng (\texttt{Mengzhe.Geng@nrc-cnrc.gc.ca}) and Xu Li (\texttt{xuliustc1306@gmail.com}).} \quad Max W. Y. Lam$^{2}$ \quad Yuewen Cao$^{2}$ \\
Shoukang Hu$^{2}$ \quad Jing Xu$^{2}$ \quad Marc Tessier$^{1}$ \quad Xu Li$^{2}$\footnotemark[1] \\[5pt]
{\normalfont\small $^1$ National Research Council Canada \quad $^2$ The Chinese University of Hong Kong} \\
\end{tabular}
}
\date{}

\begin{document}
\pagestyle{plain}
\raggedbottom
\setlength{\parskip}{0.3pc}
\setlength{\belowcaptionskip}{4pt}
\maketitle

\begin{abstract}
Removing audio can shift a visual predictor's output without showing whether
the task uses the intended audio information. We introduce Conditional Input
and Visual Interpretation Checks (CIViC), a four-gate protocol that links an
interpretation to an intervention, held-out task comparison, or endpoint
control. We demonstrate CIViC on actor-disjoint face-raster tests, a synthetic
audio-feature mixture, and source-video-disjoint SpeakerVid proxy tasks. In the
static face-raster test, cross-actor audio replacement meets the registered
detection rule with a small effect; on a separate temporal endpoint,
constant-audio fits score better than audio-trained fits in four matched
seeds. In the synthetic mixture, the condition-augmented fit has higher
1,000-step RMSE than the audio-only fit (0.1807 vs. 0.1707), with the gap
shrinking during training. In the SpeakerVid proxy, the five-feature-zero fit (five low-level audio
summaries zeroed; identity, dialogue, and condition inputs retained) has lower
whole-frame grayscale RMSE than aligned audio across four held-out
conditions and five paired seeds, while aligned audio has lower segment-drift
error. The results show that output activation, held-out error, and endpoint
controls support different interpretations across these proxy tasks.
\end{abstract}

\section{Introduction}

Audio-conditioned video and avatar models often expose paths for speech,
speaker identity, timing, and structured scene information
\citep{corona2025vlogger,kong2025multitalk,anytalker2025,peng2025dualtalk,
chu2026unils}. A common intervention removes or perturbs one path and checks
whether the output changes. That test is useful, but it leaves several
interpretations open. A model can react to audio removal while ignoring the
identity of replacement audio. An auxiliary condition can alter predictions
while worsening the held-out task. A proxy ordering can appear favorable even
when a condition-only comparator predicts its target better than the
audio-informed model.

This paper asks a narrow question: which checks distinguish input activation
from a supported audio-conditioned visual prediction statement in these controlled
tasks? CIViC answers this question with four gates. Each gate targets one alternative
explanation. We fix the split, intervention, score, aggregation rule, and
decision rule before held-out scoring. CIViC is not a video generator. It is a
diagnostic framework that makes a conditional interpretation falsifiable before
it is used to motivate a broader audiovisual statement. It does not establish that
these four checks are necessary or sufficient outside the tested settings.

Prior audio-driven avatar systems evaluate generation, synchronization, or
multi-person control~\citep{corona2025vlogger,kong2025multitalk,anytalker2025,
peng2025dualtalk,chu2026unils}; other work studies disentangled audio
representations or disrupts audio control through portrait perturbations
\citep{mittal2020disentangled,gan2025silence}. MultiTalk targets audio--person
binding, and DualTalk studies speaker--listener role transitions. CIViC does not
rank those generators or assess their perceptual outputs. Its contribution is a
mapping from a specified conditional interpretation to the matched intervention,
aggregation unit, and endpoint needed to evaluate it; the case studies introduce
no new architecture.

The closest intervention protocol is the lip-leakage evaluation of
\citet{yaman2026identity}, which tests silent input, mismatched audio, matched
audio, and reference-frame selection on talking-face generators using LRS2.
These tests already show why matched synthesis alone cannot identify audio
control; our removal and replacement actions are not new. CIViC adds a held-out
task comparison and a condition-only endpoint comparator within a registered
interpretation rule. It does not evaluate the generators or perceptual endpoints
in that work. Weng et al. separate speech, effects, and music into temporal
controls~\citep{weng2025mtv}; our feature-space study instead tests a withheld
combination of synthetic perturbations and a factor vector against a compact
proxy. Across these papers, the tasks and evidence differ; CIViC is a bounded
interpretation-to-control mapping, not a new identification method or a general
evaluation standard.

Table~\ref{tab:practice-comparison} makes the distinction from familiar
evaluation actions explicit. Audio removal, replacement, held-out scoring, and
shortcut controls each answer a useful but narrower question. CIViC does not
replace any of them; it registers the interpretation for which each is needed
and blocks that interpretation when its required evidence fails.

\begin{table}[H]
\centering
\small
\caption{From common evaluation actions to bounded interpretations. The entries
are complementary, not a ranking of prior work.}
\label{tab:practice-comparison}
\begin{tabularx}{\linewidth}{@{}l Y Y@{}}
\toprule
\textbf{Evaluation action} & \textbf{What it can establish} & \textbf{What CIViC adds before the stated interpretation} \\
\midrule
Audio removal or ablation & The registered path affects this output & A matched replacement for a value-specific statement \\
Input replacement & The stated alternative changes this endpoint & Held-out task evidence when benefit is asserted \\
Held-out task score & An ordering on the registered endpoint & A shortcut control for a transfer or proxy statement \\
Perturbation or defense test & Robustness to the tested disruption & A gate-specific stop rule and retained failed gate \\
\bottomrule
\end{tabularx}
\end{table}

We provide the portable evaluation toolkit and lightweight checks in the
public GitHub repository\footnote{\url{https://github.com/MENGZHEGENG/Audio-to-Video}}. It includes deterministic
fixtures and a pinned lightweight environment, but not restricted study inputs
or assets needed to replay the full experiments.

We apply CIViC to actor-disjoint face-raster checks, a synthetic feature-mixture
task, and a source-video-disjoint SpeakerVid proxy study~\citep{wang2026speakervid}.
The studies retain a small static detection, an early proxy disadvantage that
shrinks with fitting, and a source-video-disjoint endpoint warning. They do not
establish general audio-conditioning benefit or failure. Disjointness is a split
property: repeated seeds and hardware runs do not add independent actors or
source videos.

\textbf{Contributions.} The paper makes four contributions:
\begin{enumerate}
    \item CIViC, a four-question diagnostic framework that separates input activation,
    substitution specificity, held-out benefit, and endpoint validity;
    \item actor-disjoint CREMA-D raster checks: the static protocol satisfies its
    registered detection rule but has a very small cross-actor audio effect,
    while a separate temporal renderer does not establish an audio benefit on
    its endpoint;
    \item a synthetic audio-feature study over 205/24/27 source clips,
    with zero cross-split source-clip, original-video-group, video-namespaced-speaker,
    and exact-target overlap, showing an active condition branch, a shrinking error
    gap with training, policy-dependent controls, and a condition-only warning
    on a separate scalar endpoint; and
    \item a new source-video-disjoint SpeakerVid proxy study with a five-arm,
    five-seed control matrix, four-platform execution repeats, and a post hoc
    check across its four held-out source-video groups.
\end{enumerate}

Table~\ref{tab:gate-rules} records the decision rule. A detection excludes only
its named alternative; broader or practically important interpretations also
need a task-scale threshold or equivalence region registered with its source,
unit, direction, and failure action. No threshold was registered for the tiny
CREMA-D cross-actor contrast, so it remains practically unclassified. The
three-seed Study~II result is a fixed-run direction without a held-out-case
interval or population inference; zero cross-split overlap establishes split
provenance, not a population interval.

\begin{table}[H]
\centering
\small
\caption{Interpretation rules. A detection result excludes only the named alternative; practical use needs a task-scale rule.}
\label{tab:gate-rules}
\begin{tabularx}{\linewidth}{@{}l Y Y@{}}
\toprule
\textbf{Gate} & \textbf{Required evidence} & \textbf{What a detection does not establish} \\
\midrule
Activation & Output changes after the registered removal & Which input values the model uses \\
Substitution & Matched replacement changes the stated endpoint & Task benefit or general audio use \\
Benefit & Matched held-out error comparison & A mechanism for any observed gap \\
Endpoint validity & Control excludes the named shortcut & Validity outside the checked endpoint \\
\bottomrule
\end{tabularx}
\end{table}

\section{CIViC maps interpretations to four evidence questions}
\label{sec:protocol}

\begin{figure}[H]
\centering
\includegraphics[width=\textwidth]{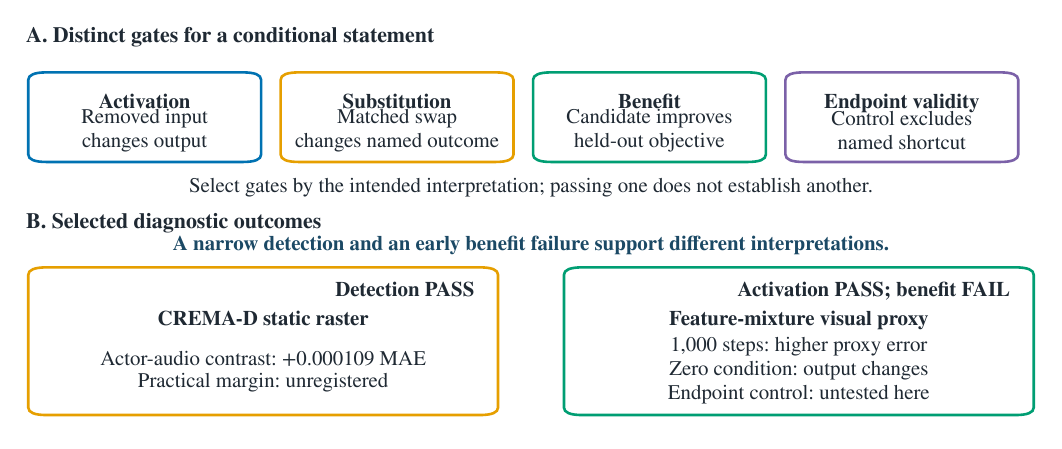}
\caption{Four interpretation gates (A) and two selected diagnostic summaries
(B). CREMA-D meets its registered fixed-pair detection rule but has no practical
threshold. In the synthetic feature-mixture proxy, the condition changes
predictions while early held-out error is higher; that gap shrinks with
training. Its matched endpoint control is untested. The newer SpeakerVid proxy
comparison is described in Section~\ref{sec:speakervid-study}, not shown
here. Color and text distinguish the gates.}
\label{fig:gates}
\end{figure}

Let $f(x,a,z)$ map a fixed visual context $x$, audio $a$, and optional
auxiliary condition $z$ to a motion target $y$. A nonzero difference between
$f(x,a,z)$ and $f(x,0,z)$ only shows that removing audio changes the output.
Figure~\ref{fig:gates} maps four CIViC gates to the alternative explanation
each gate excludes. The gates are not a universal checklist. A study runs the
gates required by its intended interpretation. For example, a speaker-specific
statement needs a registered speaker-audio substitution, a mixed-condition
benefit needs a matched task comparison, and a proxy-transfer statement needs
an endpoint control that excludes non-audio shortcuts.

The gates are selected for the intended interpretation and fix the split,
intervention, score, aggregation unit, and decision rule before held-out scoring.
An endpoint warning excludes the named interpretation but does not identify its
shortcut; a new control must exclude the competing endpoint result before that
interpretation is restored. Repeated fits measure run variation, not added
population coverage. For a practically useful actor-matched raster statement,
the protocol card therefore requires an audio replacement, a held-out score, a
task-scale threshold, and an endpoint control when the target is a proxy.
Figure~\ref{fig:gates} contrasts the narrow CREMA-D detection with the failed
early-benefit check; Table~\ref{tab:crema} gives the paired raster results.

\section{Study I: static and temporal face-raster checks}
\label{sec:raster}

\subsection{CREMA-D meets a narrow fixed-pair detection rule}

The first raster application uses CREMA-D speech-video recordings
\citep{cao2014cremad}. A fixed protocol assigns 73 actors to development, 9 to
validation, and 9 to sealed confirmation. The renderer receives an identity
raster decoded at 0.25 seconds and a waveform-derived log-mel window from 0.75
to 1.25 seconds. It predicts the $64\times64$ target raster at 1.0 seconds.
The audio window extends 0.25 seconds beyond the target time; this is a
centered-window raster reconstruction.
Inputs exclude emotion, intensity, sentence, and demographic labels.
For this raster application, mean absolute error (MAE) averages absolute RGB
intensity differences over all $64\times64\times3$ values after division by
255, so it is measured on a $[0,1]$ channel scale. We set the confirmation
protocol, pairings, and bootstrap rule before evaluating the 732 confirmation
clips. For actor-audio replacement, actors follow a fixed cycle, and a
deterministic mapping from protocol and clip identity selects a clip from the
next actor.
Utterance, emotion, and intensity are not matched, so this control changes
speaker and speech content together.

Table~\ref{tab:crema} reports the fixed paired contrasts. Every registered
lower interval bound is positive. The cross-actor-audio effect is 0.000109
MAE, however. This meets the registered sign-based detection rule, but no
task-scale threshold was registered, so its practical relevance is
unclassified. It does not establish robust speaker transfer, temporal
generation, natural multi-person behavior, active-speaker accuracy, or
human-perceptual quality.

\begin{table}[H]
\centering
\small
\caption{CREMA-D static confirmation on 732 fixed pairs from an actor-disjoint
split. Differences are control MAE minus aligned MAE; aligned MAE is 0.02907.
Paired intervals use 10,000 clip-level resamples with seed 1701 for the fixed
checkpoint and pair set; they do not quantify actor-population uncertainty.
Positive differences favor aligned input.}
\label{tab:crema}
\begin{tabularx}{\linewidth}{@{}l Y Y@{}}
\toprule
\textbf{Control} & \textbf{Control MAE $-$ aligned MAE $\uparrow$} & \textbf{Paired 95\% interval} \\
\midrule
Zero audio & +0.131475 & [0.127163, 0.135640] \\
Actor-audio replacement & +0.000109 & [0.000034, 0.000186] \\
Identity-raster replacement & +0.064860 & [0.062795, 0.066897] \\
\bottomrule
\end{tabularx}
\end{table}

\subsection{A separate temporal renderer fails its audio controls}

The static test does not answer whether audio improves a sequence renderer. A
separately registered renderer was fitted on 5,038 development sequences,
selected on 689 validation sequences, and evaluated on 708 sealed sequences.
Aligned MAE was 0.030449; shifting audio raised it by 0.0000176, while
repeating the identity raster lowered it by 0.010592. Without an interval or
practical margin for the shift, this check does not establish a sequence-level
audio benefit or measure perceptual video quality.

\subsection{A separate confirmation fails the audio-replacement gate}

A second actor-disjoint raster protocol uses the Ryerson Audio-Visual Database
of Emotional Speech and Song (RAVDESS)~\citep{livingstone2018ravdess}. An
earlier result summary scored 1.0 seconds rather than the registered
2.0-second target, so we evaluated the 240-pair protocol and checkpoint at
2.0 seconds. Aligned MAE is
0.08990. Zero audio raises it by 0.20931 (paired 95\% interval
$[0.20469,0.21358]$), while another actor's audio changes it by $-0.00045$
($[-0.00129,0.00039]$); the registered replacement rule fails. Exploratory
filename-matched cross-actor and same-actor statement swaps change MAE by
$-0.00102$ ($[-0.00491,0.00141]$, four actors) and $+0.00004$
($[-0.00017,0.00025]$), respectively.

A same-architecture constant-audio fit scores 0.08652 on those targets,
versus 0.08990 for the audio-trained fit (difference $-0.00338$,
descriptive four-actor bootstrap 95\% interval $[-0.00512,-0.00164]$). Three matched-seed fits
also favor constant audio; all four differences range from $-0.00338$ to
$-0.00216$. A fresh four-seed, 80-epoch repeat favors constant audio in all four pairs, but
trajectories diverge before epoch 40, so it does not isolate extra fitting.
These retrospective comparisons do not create a registered benefit gate.
Removal is activation-sensitive, but replacement fails and neither swap
isolates speaker identity because speech content also changes. Appendix
\ref{app:ravdess-integrity} gives the registered endpoint and within-trajectory
details.

\subsection{An incompatible external audit tests a separate boundary}

We apply the frozen renderer to 33 AVA Active Speaker validation recordings
\citep{roth2020avaactive}. The shared soundtrack has no per-face audio, so this
is an external boundary check only. Aligned, shifted, and zero-audio AUROC are
0.63678, 0.63670, and 0.63672; the aligned-minus-shifted interval is
$[-0.00017,0.00028]$. This check supplies no speaker-attribution evidence.

\section{Study II: an active condition offers no stable benefit on a withheld feature mixture}
\label{sec:mixed}

\subsection{Controlled composition protocol}

Study~II uses 819 indexed synthetic feature mixtures from a private,
provenance-approved source family: 615 train, 96 validation, and 108 test
records expanded from 205/24/27 source clips. The train, validation, and test
splits have zero pairwise overlap in source clips, original-video groups,
video-namespaced speakers, and exact target face arrays. These are provenance
and split checks, not independent-sample or population claims. Each mixture starts with five speech summaries and adds the same nuisance
sequence to both speakers. Environment-like features are seeded normalized noise;
music-like features are seeded sinusoids, with either slow/transient noise and an
optional beat envelope. They are five-coordinate feature arrays, not waveforms,
scaled to 0.25 and 0.20 times speech-feature standard deviation.

The source feature index lists 5-dimensional audio, 6-dimensional dialogue,
128-dimensional face, and 256-dimensional identity streams, all at 25 Hz. The
128-coordinate target is copied from the source speaker-axis pack. Its extractor
decodes whole frames at 25 Hz with \texttt{ffmpeg\_gray\_proxy}, scales each frame to
16-by-16 grayscale, averages adjacent vertical pixel pairs to obtain an 8-by-16
array, and subtracts 0.5. Speaker-axis construction gates the face target with
$0.20+0.80a$, where $a$ is annotation-derived activity; MixText copies this
target into each condition. Streams are paired by frame index and truncated to
their common frame count, with no presentation-timestamp sidecar retained. The
identity proxy is the mean of the first eight grayscale frames repeated over
speaker slots. RMSE is computed over these dimensionless coordinates without
per-condition rescaling; it is not a physical motion or video-quality measure.

Both arms receive five audio summaries, this image-derived 256-coordinate
identity proxy, and six annotation-derived dialogue coordinates. The splits
have zero overlap in source clips, original-video groups, and video-namespaced
speakers, but real-person identity disjointness remains unverified. The added
condition is the seven-bit vector
$[s,e,m,e_0,e_1,m_0,m_1]$, encoding speech, environment, music, and two type
indicators per nuisance stream. The shared past-only backbone uses a 96-frame
context, width 256, and four attention heads. Adding the condition increases
input width; the joint mixed condition is withheld from training. The primary
comparison is the 1,000-step checkpoint, with the same three seeds checked at
3,000 and 10,000 steps. Separate analyses test training with all conditions
present, 24-seed training-support policies, five-seed parameter-matched
controls, and source-disjoint scalar endpoint shortcuts; their estimands are
not pooled. The registered fixed-run rule requires positive structured-minus-
audio gaps for all three seeds to indicate no benefit at 1,000 steps. This is a
directional result without a held-out-case interval or population inference.

\subsection{The early error gap shrinks with training}

\begin{figure}[H]
\centering
\includegraphics[width=\textwidth]{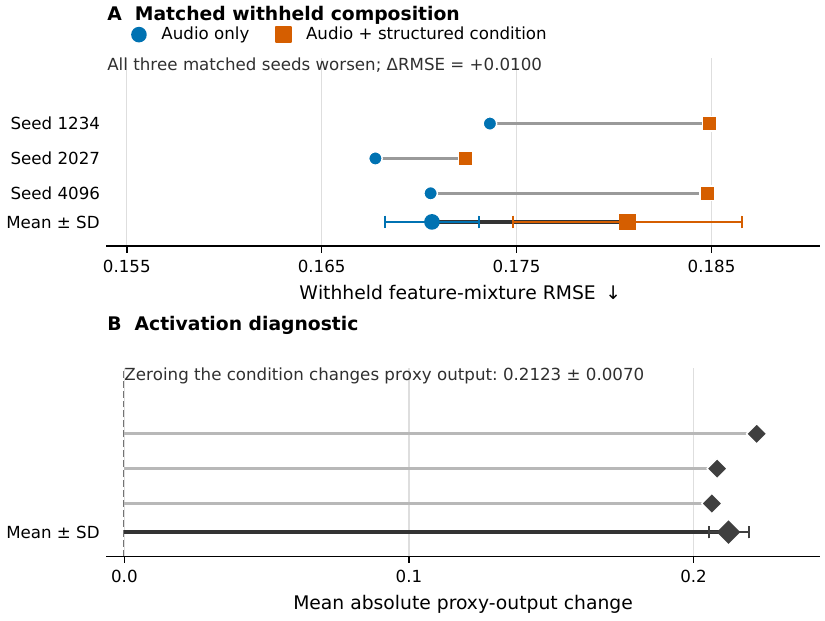}
\caption{Registered 1,000-step feature-mixture evidence. (A) Adding the
condition raises withheld-composition RMSE in all three matched seeds; the
target is a dimensionless stored proxy. (B) Zeroing the condition changes
proxy predictions in the same runs. Points are seed values; markers are mean
$\pm$ standard deviation. Activation is not a task-error comparison.}
\label{fig:mixed-results}
\end{figure}

Table~\ref{tab:mixed} reports three checkpoints from the same runs. At 1,000
steps, all three paired mixed-condition gaps are positive; by 10,000 steps the
mean is $0.0019\pm0.0032$ and one seed gap is negative. The early disadvantage
therefore shrinks with fitting. Figure~\ref{fig:mixed-results} shows that
zeroing the condition changes predictions by $0.2123\pm0.0070$ and replacing
the mixed vector with the clean vector changes them by $0.1473$ on average;
neither displacement measures task benefit. Clean, environment-only, and
music-only test gaps are $-0.000284$, $0.001008$, and $0.000909$, versus
+$0.010032$ on the withheld mixture. Relative to clean, mixture error rises by
+$0.0095$ for audio only and $0.0198$ with the condition. The 24-seed support
audit has a positive structured-minus-audio gap for every policy
(Table~\ref{tab:policy-audit}). With all conditions in training, a separate
three-seed control gives mixed RMSE $0.1659\pm0.0027$ for audio only and
$0.1684\pm0.0047$ with the condition; it does not test withheld composition.

\begin{table}[H]
\centering
\small
\caption{Twenty-four-seed, 1,000-step training-support audit on the held-out mixed
condition (mean $\pm$ population standard deviation). The gap is
added-condition minus audio-only RMSE; positive gaps favor audio only. Each
policy changes training support and is not pooled with the three-seed result.}
\label{tab:policy-audit}
\vspace{2pt}
\begin{tabularx}{\linewidth}{@{}l Y Y Y@{}}
\toprule
\textbf{Training support} & \textbf{Audio-only RMSE $\downarrow$} & \shortstack[l]{\textbf{Added-condition}\\\textbf{RMSE $\downarrow$}} & \textbf{Gap $\downarrow$} \\
\midrule
Single clean & $0.4249\pm0.0097$ & $0.6921\pm0.0218$ & $+0.2672\pm0.0276$ \\
Single environment & $0.1638\pm0.0026$ & $0.4192\pm0.0239$ & $+0.2554\pm0.0240$ \\
Single music & $0.1681\pm0.0025$ & $0.4169\pm0.0247$ & $+0.2487\pm0.0252$ \\
Clean plus environment & $0.1685\pm0.0027$ & $0.4121\pm0.0225$ & $+0.2436\pm0.0231$ \\
\bottomrule
\end{tabularx}
\end{table}

\begin{table}[H]
\centering
\small
\caption{Withheld mixed-composition error at three checkpoints of the same
three-seed runs (mean $\pm$ population standard deviation; lower RMSE is better).
Audio only omits the seven-bit condition; both arms retain speaker and dialogue
inputs. The gap is added-condition minus audio-only RMSE; positive favors audio only.}
\label{tab:mixed}
\vspace{2pt}
\begin{tabularx}{\linewidth}{@{}l Y Y Y@{}}
\toprule
\textbf{Steps} & \textbf{Audio-only RMSE $\downarrow$} & \shortstack[l]{\textbf{Added-condition}\\\textbf{RMSE $\downarrow$}} & \textbf{Gap $\downarrow$} \\
\midrule
\textbf{1,000} & $0.1707\pm0.0024$ & $0.1807\pm0.0059$ & $+0.0100\pm0.0040$ \\
3,000 & $0.1338\pm0.0004$ & $0.1366\pm0.0023$ & $+0.0028\pm0.0019$ \\
10,000 & $0.1237\pm0.0014$ & $0.1256\pm0.0042$ & $+0.0019\pm0.0032$ \\
\bottomrule
\end{tabularx}
\end{table}

\subsection{A parameter-matched audit exposes a training-support mismatch}

The primary comparison changes input width, so we separately evaluated three
parameter-matched control policies across five seeds under clean-plus-environment
support. Each uses the
seven-coordinate interface and 4,908,416 parameters (Table~\ref{tab:alignment}).
The dummy-condition model scores $0.1822\pm0.0025$; training with shuffled
conditions but aligned evaluation scores $0.4248\pm0.0217$; and five-feature-zero with
aligned conditions scores $0.5003\pm0.0278$. These policies change multiple
inputs or train--evaluation support, so their gaps isolate neither audio nor
condition alignment. The same-checkpoint intervention in
Figure~\ref{fig:mixed-results} supplies the condition-activity test. These
controls neither make the primary three-seed comparison parameter matched nor
show benefit. We exclude a train-by-evaluation comparison because its
underlying results could not be verified.

\begin{table}[H]
\centering
\small
\caption{Five-seed parameter-matched controls on the held-out mixture under
clean plus environment training support (mean $\pm$ population standard
deviation; seeds 1234, 2027, 4096, 8191, 16384). All entries use seven condition
coordinates and 4,908,416 parameters. Shuffled values are used in training
but aligned values in evaluation; this is a policy-specific diagnostic.}
\label{tab:alignment}
\begin{tabularx}{\linewidth}{@{}l Y Y Y@{}}
\toprule
\textbf{Control} & \textbf{Audio} & \textbf{Condition relation} & \textbf{Mixed RMSE $\downarrow$} \\
\midrule
Dummy condition & Observed & Deterministic dummy & $0.1822\pm0.0025$ \\
Shuffled condition & Observed & Train shuffled; test aligned & $0.4248\pm0.0217$ \\
Five-feature-zeroed & Zeroed & Aligned & $0.5003\pm0.0278$ \\
\bottomrule
\end{tabularx}
\end{table}

\subsection{A condition-only comparator raises an endpoint warning}
\label{sec:endpoint-warning}

A source-file-disjoint CREMA-D ridge check (71 clips, three held-out actors,
five mixtures) gives RMSE $0.00190136$ for the interaction model versus
$0.00189705$ for audio only; no paired interval or practical margin was
retained. A separate RAVDESS ridge check over 1,441 clips and 24 source groups
gives structured-minus-audio RMSE $-0.00000924$ with interval
$[-0.00002350,0.00000148]$; metadata-only RMSE ($0.00881499$) is below both
audio-informed variants ($0.00926477$ and $0.00927402$). Neither check
isolates audio use in the scalar proxy.

A separate source-disjoint neural follow-up fits fixed width-32 ReLU regressors
on CREMA-D and evaluates four held-out GRID talkers. Its structured-minus-audio
contrast is $-0.0023303$ with talker-clustered interval
$[-0.0032104,-0.0006693]$, but a condition-only comparator reaches RMSE near
$0.00109$, below the structured ($0.00865$--$0.01341$) and audio-only
($0.01231$--$0.01592$) models. These descriptive scalar-proxy checks show that
an apparently favorable ordering cannot support the intended audio-conditioning
interpretation; they do not identify the shortcut in Study~II's
feature-mixture comparison.

\section{Study III: a source-video-disjoint SpeakerVid proxy study}
\label{sec:speakervid-study}

Study~II's primary target also has 128 coordinates, but its separate shortcut
follow-ups use a scalar proxy. We run a new source-video-disjoint study on
SpeakerVid-5M~\citep{wang2026speakervid}. The study uses 865 decoded two-speaker
windows from 31 videos (649/108/108 train/validation/test records across
23/4/4 videos). Source-clip, source-video, and exact-target overlap are zero
between splits. One unverified video was excluded; this is a new source pack,
not a replay of Study~II. It supplies no matched comparator for that study.

Each record has five low-level audio summaries, an image-derived identity
descriptor, annotation-derived dialogue features, and a 128-coordinate visual
target. The five audio summaries are log energy, root-mean-square amplitude,
zero-crossing rate, mean, and standard deviation, computed on 40-ms windows
with 40-ms hops (25 features per second). The video extractor samples
whole frames at a nominal 25 Hz, scales each to 16-by-16 grayscale, averages
adjacent vertical pixel pairs, and subtracts 0.5 to form an 8-by-16 (128-value)
raster. It uses the whole frame, not a face or mouth crop, landmarks, or a
semantic motion representation. The feature files retain the nominal frame
rate but no presentation timestamps, so streams are paired by frame index and
truncated to their common frame count. The six dialogue coordinates are two
speaker-activity indicators, active-speaker count, turn-start, turn-end, and
overlap, derived from supplied annotations. The 256-value identity input is the
mean 16-by-16 grayscale frame over the first eight frames, repeated over time
and speaker slots; it is not a pretrained speaker embedding. In the
speaker-axis construction, audio is activity-weighted and the visual target is
scaled by a gate of $0.20+0.80a$, where $a$ is annotation-derived speaker
activity. The target and remaining dialogue input therefore share activity
information. This endpoint does not establish mouth motion or audio-video
synchrony. A deterministic feature-mixture generator provides
clean, environment-like, and music-like training conditions plus a held-out
mixture; its seven-dimensional vector encodes synthetic presence/type, not a
waveform or natural-language prompt. Five trained arms are aligned audio
summaries and condition, five-feature-zero with aligned condition, observed
summaries with the condition branch omitted, observed summaries with a dummy
condition, and observed summaries with batch-shuffled conditions. They share
the data, targets, split,
1,000-step schedule, and five seeds (1234, 2027, 4096, 8191, 16384).

The five-feature-zero arm sets the five audio-summary inputs to zero while
retaining the image-derived identity descriptor, annotation-derived dialogue
features, and aligned condition vector. Because that condition vector encodes
synthetic speech presence and type, the arm is not a silent-input condition and
does not remove all audio-derived information. We repeated the
full matrix on four accelerator platforms: one H100, two A100
(labeled A100-1 and A100-2 in Table~\ref{tab:speakervid-platform-runs}), and
one V100. The 100 fits comprise 25 fits per platform and reuse one source pack
across four platforms; they are compute replications, not new independent
evidence. A
post-hoc evaluation groups aligned-audio and five-feature-zero predictions by the four
held-out videos; it describes those groups only.

\subsection{Five-feature-zero has lower test error across platforms and source groups}

Table~\ref{tab:speakervid-platform-runs} reports the paired test difference in
root mean squared error (RMSE) over the 128 grayscale-raster coordinates,
computed as control minus aligned audio for each seed. Negative values favor
the control. Five-feature-zero has a negative
mean in all 16 platform-by-condition cells, and all five paired seeds are
negative in each cell. On validation, the same contrast is negative for all five
seeds in every condition on every platform.

This ranking is specific to raster-reconstruction RMSE. A secondary segment-drift proxy,
for which lower is better, has the opposite mean ordering for five-feature-zero
minus aligned audio in all 16 test platform-condition cells. The four-platform
ranges of five-seed means are $[0.0057,0.0064]$ for clean,
$[0.0076,0.0079]$ for environment-like, $[0.0069,0.0070]$ for music-like,
and $[0.0043,0.0050]$ for mixed conditions. These are repeated fits on one
source pack, not intervals over new videos. The evaluator flattens the
predicted feature windows, divides that array into 10-second segments, and
compares later segment means with the first; it does not track drift in a
continuous rendered clip. Appendix~\ref{app:speakervid-detail},
Table~\ref{tab:speakervid-drift}, reports the contrasting proxy alongside
raster-reconstruction RMSE and its seed-level signs. Neither proxy establishes rendered-video
quality; the drift result prevents an unqualified statement that five-feature-zero is better.

\begin{table}[H]
\centering
\small
\setlength{\tabcolsep}{4pt}
\caption{Paired test raster-reconstruction RMSE differences for the five-feature-zero control,
computed as five-feature-zero minus aligned audio. Entries are mean $\pm$ sample
standard deviation over five paired seed differences. Negative values favor
five-feature-zero. The four platform runs share one source pack and are hardware or
execution replications, not independent data. The secondary segment-drift proxy
favors aligned audio in every condition; Table~\ref{tab:speakervid-drift}
reports that opposing result.}
\label{tab:speakervid-platform-runs}
\begin{tabularx}{\linewidth}{@{}l *{4}{>{\centering\arraybackslash}X}@{}}
\toprule
\textbf{Condition} & \textbf{H100 $\Delta$RMSE$\downarrow$} &
\textbf{A100-1 $\Delta$RMSE$\downarrow$} &
\textbf{A100-2 $\Delta$RMSE$\downarrow$} &
\textbf{V100 $\Delta$RMSE$\downarrow$} \\
\midrule
Clean & $-0.0142\pm0.0036$ & $-0.0143\pm0.0035$ & $-0.0143\pm0.0035$ & $-0.0142\pm0.0036$ \\
Environment-like & $-0.0214\pm0.0022$ & $-0.0215\pm0.0022$ & $-0.0215\pm0.0022$ & $-0.0216\pm0.0024$ \\
Music-like & $-0.0210\pm0.0019$ & $-0.0211\pm0.0019$ & $-0.0211\pm0.0019$ & $-0.0213\pm0.0020$ \\
Mixed & $-0.0206\pm0.0024$ & $-0.0208\pm0.0024$ & $-0.0208\pm0.0024$ & $-0.0212\pm0.0027$ \\
\bottomrule
\end{tabularx}
\end{table}

The audio-only model, which keeps observed speech but omits the condition
branch, has lower mean test RMSE in all 16 platform-condition cells. All five
seed differences are negative except music, where three platforms have four
negative seeds and V100 has five. Because input dimensionality changes, this is
a branch-omission comparison, not an isolated audio effect. The dummy-condition
model has higher RMSE in every cell and seed. Shuffled conditions have small
positive test means ($0.0006$--$0.0022$) but variable seed signs. On validation,
audio-only means are lower in all cells (four of five negative seeds for clean,
all five for the other conditions); dummy-condition means are higher in every
cell and seed. Shuffled-condition means are positive in all 16 validation cells,
but seed signs vary. Validation remains separate from the test result.
Appendix~\ref{app:speakervid-detail}, Table~\ref{tab:speakervid-controls},
shows the full test-condition contrast: the audio-only arm favors omission of
the condition branch. The dummy-condition arm has higher raster-reconstruction
RMSE, while the shuffled-condition arm has small positive mean differences and
variable seed signs. Their secondary segment-drift means have a different ordering, detailed
in Appendix~\ref{app:speakervid-detail}; this is not a single-metric ranking
of the complete system.

The post-hoc grouping has 108 unique test windows from four videos, each
scored in four conditions (432 video-condition rows). Every video-condition
mean and all five seed differences are negative on each platform; means range
from $-0.0334$ to $-0.0123$. These four groups are the complete held-out video
subset, so this is descriptive consistency, not a population interval.
Appendix~\ref{app:speakervid-detail}, Table~\ref{tab:speakervid-units},
separates the four source videos; no one video reverses the five-feature-zero ordering
in any condition, but these are not four independent test populations.

On test, replacing the aligned condition vector with zero moves predictions by $0.2127$--$0.2129$ (seed SD $0.0115$--$0.0117$) in
aligned fits and $0.2614$--$0.2619$ (SD $0.0104$--$0.0105$) in five-feature-zero fits.
Replacing it with the clean vector moves them by $0.1570$ (SD $0.0105$) and
$0.1338$--$0.1341$ (SD $0.0080$), respectively. These are displacements, not
condition correctness or task benefit; text fields are absent for audio-only
because it has no condition branch.

Raster-reconstruction RMSE is the primary endpoint. Active/inactive RMSE are components;
causal selectivity, target fidelity, and listener responsiveness are derived;
non-target leakage duplicates inactive RMSE; and speaker-order delta is a raw
perturbation diagnostic. Identity stability measures input-order invariance,
not identity fidelity, while the coarse CIViC score is uncalibrated. For
five-feature-zero fits, audio-video sync and modality consistency use a zero-energy
fallback of 0.5, and lag is at its $-400$ ms search boundary; none is interpreted
as synchronization performance. Turn-timing error remains a secondary proxy
output. The bounded result is an endpoint-specific ordering for this compact
target: the separately trained five-feature-zero fit has lower
raster-reconstruction RMSE, while aligned audio has lower segment-drift proxy
error on these held-out records. This comparison establishes no natural-video
quality or validity for Study~II's separately sourced target.

The Study~III target contains 299,212 training and 50,788 test frame--speaker
observations from 23 and four videos, respectively. The median of the 128
coordinate-wise population standard deviations was 0.2047 (train) and 0.3012
(test), computed over clean observations with the observation count as divisor.
Test adjacent-frame RMS was 0.0198, the square root of mean squared differences
between consecutive frame indices within records, pooled over both speaker
slots and all coordinates. A coordinate-wise mean fitted on clean training
frames scored RMSE 0.3477 over 424 clean test windows using the model's
96-frame window, 48-frame stride, final-window rule, and coordinate-wise
aggregation (10,420,224 values). Five-seed clean-test mean RMSE was 0.2005 for
aligned audio and 0.1864 for five-feature-zero. Both fitted predictors improve
on this constant-mean baseline for the raster endpoint; this does not isolate
audio, since five-feature-zero retains identity, dialogue, and aligned-condition
inputs. The pooled target mean also shifts from $-0.1526$ in training to
$0.0114$ in test, so this baseline comparison is descriptive under a changed
target distribution, not a measure of audio-specific predictability.
These dimensionless statistics show raster variation, not mouth motion or
audio-predictable dynamics: without source media or presentation timestamps,
frame-index differences establish neither temporal motion nor audio alignment.
Video-namespaced speaker labels do not overlap across splits, but real-person
disjointness is unverified. Study~II has zero source-clip and original-video-
group overlap and uses a copied grayscale target; neither split check verifies
real-person identity or natural-video validity.

\section{Conclusion}
\label{sec:discussion}

These studies show that output activation, held-out error, and endpoint
controls support different interpretations. CREMA-D's cross-actor effect is
small and utterance-confounded under a detection rule with no practical
threshold. In Study~II, the early fit gap narrows with training, and
same-checkpoint condition replacements displace predictions without measuring
task benefit. Separate scalar proxy controls produce endpoint warnings that do
not identify a shortcut in the feature-mixture target. In Study~III, the
separately trained five-feature-zero arm has lower raster RMSE than the
aligned-audio arm, while aligned audio has lower segment-drift error. The
audio-only branch-omission fit also has lower raster RMSE, but changes input
width and therefore does not isolate a condition effect. The whole-frame
target and four-video test set support descriptive proxy findings, not
rendered-video quality, natural-mixture performance, or speaker attribution.
Human perceptual evaluation remains future work.

\section*{Reproducibility Statement}

Sections~\ref{sec:protocol}--\ref{sec:speakervid-study} specify the protocols and bounded
results. The public repository provides deterministic checks, media-free
fixtures, and a pinned lightweight environment. Full experiment replay requires
approved source material that is not hosted in the repository.

\section*{AI Use Statement}

OpenAI Codex assisted with literature search, manuscript review and editing,
and preparation of the final source and PDF packages. Gemini and Claude
reviewed wording in earlier drafts. The authors take responsibility for the
final text and citations.

{\footnotesize
\bibliographystyle{iclr2027_conference}
\bibliography{references}
}
\clearpage
\appendix
\section{SpeakerVid test contrasts by condition and source video}
\label{app:speakervid-detail}

The test endpoint here is root mean squared error over the dimensionless
128-coordinate whole-frame grayscale proxy. Each difference subtracts the aligned-audio
arm's error from the control arm's error on the same held-out records and
paired training seed. Negative differences favor the control. The four
accelerator runs reuse one source pack and the same five seeds; their spread
is descriptive hardware sensitivity, not a confidence interval over videos.

Table~\ref{tab:speakervid-drift} separates the two endpoints. Five-feature-zero
has lower raster-reconstruction error, while aligned audio has lower error on
the flattened-window segment-drift proxy in all four conditions. Positive
paired drift differences occur in four of five seeds for clean and mixed
conditions, and all five seeds for environment-like and music-like conditions.
These ranges summarize four platforms on the same videos and are not
independent-video intervals; flattened windows do not measure continuous-video
quality.

\begin{table}[H]
\centering
\small
\caption{Five-feature-zero minus aligned-audio test differences for two lower-is-better
proxies. Negative raster-reconstruction RMSE differences favor five-feature-zero,
while positive drift differences favor aligned audio. The flattened-window drift
statistic does not measure continuous-video quality. Ranges span four platform
means computed from the same five seeds and four-video test subset; they are
descriptive, not confidence intervals.}
\label{tab:speakervid-drift}
\vspace{4pt}
\begin{tabularx}{\linewidth}{@{}l *{3}{>{\centering\arraybackslash}X}@{}}
\toprule
\textbf{Condition} & \textbf{Raster $\Delta$RMSE $\downarrow$} &
\textbf{Drift $\Delta$ $\downarrow$} &
\textbf{Positive drift seeds} \\
\midrule
Clean & $[-0.0143,-0.0142]$ & $[0.0057,0.0064]$ & $4/5$ \\
Environment-like & $[-0.0216,-0.0214]$ & $[0.0076,0.0079]$ & $5/5$ \\
Music-like & $[-0.0213,-0.0210]$ & $[0.0069,0.0070]$ & $5/5$ \\
Mixed & $[-0.0212,-0.0206]$ & $[0.0043,0.0050]$ & $4/5$ \\
\bottomrule
\end{tabularx}
\end{table}

Table~\ref{tab:speakervid-controls} separates four controls that change
different things. Zeroing the five audio summaries during fitting has lower
error in every condition, and removing the condition branch also lowers the
mean error.
The dummy condition raises error for every seed. Shuffling condition vectors
raises the mean error slightly, but the seed-level direction varies. These
comparisons cannot be collapsed into one estimate of audio or condition use:
the controls differ in fitted inputs, branch capacity, or condition pairing.
The secondary segment-drift proxy adds a further qualification. Relative to
aligned audio, audio-only lowers both mean raster-reconstruction RMSE and mean drift in every
platform-condition cell. Dummy and shuffled conditions raise mean raster-reconstruction RMSE
but lower mean drift in every such cell. For example, across the four platforms
in the mixed condition, dummy-minus-aligned drift ranges from $-0.0017$ to
$-0.0015$ and shuffled-minus-aligned drift from $-0.0044$ to $-0.0041$.
These are descriptive five-seed means on the same four-video test subset, not
independent-video intervals. Drift is computed over flattened feature windows,
so the opposing directions do not establish improved continuous-video quality.

\begin{table}[H]
\centering
\small
\caption{Test raster-reconstruction RMSE difference (control minus aligned audio) by
condition. Each cell is the minimum--maximum of four platform-level means,
each computed from five paired seed differences. These ranges are not
confidence intervals; platforms share the same held-out videos and seeds.
Negative values favor the named control.}
\label{tab:speakervid-controls}
\vspace{4pt}
\begin{tabularx}{\linewidth}{@{}l *{4}{>{\centering\arraybackslash}X}@{}}
\toprule
\textbf{Condition} & \textbf{Five-feature-zero $\Delta$RMSE $\downarrow$} &
\textbf{Audio-only $\Delta$RMSE $\downarrow$} &
\textbf{Dummy $\Delta$RMSE $\downarrow$} &
\textbf{Shuffled $\Delta$RMSE $\downarrow$} \\
\midrule
Clean & $[-0.0143,-0.0142]$ & $[-0.0040,-0.0037]$ & $[0.0165,0.0167]$ & $[0.0006,0.0007]$ \\
Environment-like & $[-0.0216,-0.0214]$ & $[-0.0045,-0.0042]$ & $[0.0165,0.0167]$ & $[0.0011,0.0012]$ \\
Music-like & $[-0.0213,-0.0210]$ & $[-0.0046,-0.0042]$ & $[0.0163,0.0166]$ & $[0.0011,0.0012]$ \\
Mixed & $[-0.0212,-0.0206]$ & $[-0.0106,-0.0101]$ & $[0.0108,0.0111]$ & $[0.0019,0.0022]$ \\
\bottomrule
\end{tabularx}
\end{table}

Table~\ref{tab:speakervid-units} asks whether the five-feature-zero result is
concentrated in one held-out video. The 108 unique windows come from four
videos, with 26, 20, 48, and 14 windows, respectively; each window is scored
under all four conditions. Every video-by-condition cell has a negative mean
on all four platforms, and all five paired seed differences are negative on
each platform. The stronger mixed-condition difference for Unit 2 is a
descriptive within-source observation, not an independently replicated effect.

\begin{table}[H]
\centering
\small
\caption{Five-feature-zero minus aligned-audio test raster-reconstruction RMSE within each
held-out source video. Unit numbers are labels for the four videos, not new
sampling units; $n$ counts unique windows per video, each reused across four
conditions. Cells give the minimum--maximum of four five-seed platform means,
not population intervals. Negative values favor five-feature-zero.}
\label{tab:speakervid-units}
\vspace{4pt}
\begin{tabularx}{\linewidth}{@{}lc *{4}{>{\centering\arraybackslash}X}@{}}
\toprule
\textbf{Unit} & \textbf{$n$} & \textbf{Clean $\Delta$RMSE $\downarrow$} &
\textbf{Environment $\Delta$RMSE $\downarrow$} &
\textbf{Music $\Delta$RMSE $\downarrow$} &
\textbf{Mixed $\Delta$RMSE $\downarrow$} \\
\midrule
1 & 26 & $[-0.0124,-0.0123]$ & $[-0.0175,-0.0172]$ & $[-0.0189,-0.0188]$ & $[-0.0173,-0.0169]$ \\
2 & 20 & $[-0.0135,-0.0127]$ & $[-0.0267,-0.0258]$ & $[-0.0243,-0.0237]$ & $[-0.0334,-0.0326]$ \\
3 & 48 & $[-0.0161,-0.0158]$ & $[-0.0232,-0.0231]$ & $[-0.0225,-0.0222]$ & $[-0.0204,-0.0199]$ \\
4 & 14 & $[-0.0134,-0.0132]$ & $[-0.0192,-0.0191]$ & $[-0.0199,-0.0195]$ & $[-0.0185,-0.0177]$ \\
\bottomrule
\end{tabularx}
\end{table}

Neither table estimates uncertainty over new videos. The video-level
comparison covers the complete four-video held-out subset; repeated windows,
conditions, seeds, and platforms do not increase its independent-video count.
The result warns that output sensitivity to audio or the condition vector
does not establish a benefit on this compact visual target.

\section{RAVDESS raster confirmation at the registered target}
\label{app:ravdess-integrity}

The registered RAVDESS speech-video protocol~\citep{livingstone2018ravdess}
uses an identity raster at 0.25 seconds, mono 16 kHz audio from 1.0 to 1.5
seconds, and a $64\times64$ target raster at 2.0 seconds. Sixteen actors are
assigned to development, four to validation, and four to confirmation. The
compact renderer uses a visual encoder, an audio multilayer perceptron,
audio-derived scale and bias, and a visual decoder; inputs exclude emotion,
intensity, statement, and actor identifiers. The 240 fixed-pair test
compares aligned audio with zero audio, audio from another confirmation actor,
and a swapped identity raster. Partner actors cycle through the confirmation
actors; a deterministic mapping from protocol and clip identity selects the
partner clip. Statement and emotion are not matched, so the audio replacement changes
actor and utterance together.

The registered protocol and checkpoint score the 2.0-second target on the 240
fixed pairs. Aligned MAE is 0.08990. Control-minus-aligned
contrasts are $+0.20931$ for zero audio, $-0.00045$ for actor-audio
replacement, and $+0.04654$ for identity-raster replacement. Their paired
95\% bootstrap intervals are $[0.20469,0.21358]$,
$[-0.00129,0.00039]$, and $[0.04362,0.04926]$. The registered rule
requires a strictly positive lower bound for every control; it therefore
fails. No fitting or threshold selection used
confirmation actors. The actor-audio swap also changes utterance, so it
does not identify a speaker effect.

After the registered analysis, we ran an exploratory six-condition analysis on
the same 240 unique targets and checkpoint. The cross-actor control
uses the registered partner actor but matches modality, vocal channel,
emotion, intensity, statement, and repetition in the RAVDESS filename.
Another control keeps the actor and changes only the statement code.
Neither code match guarantees synchronized phonemes in the 1.0--1.5-second
audio window. The new analysis retains each target's six MAEs and groups
uncertainty both by pair and by the four confirmation actors.
The metadata-matched cross-actor delta is $-0.00102$, with pair-bootstrap
interval $[-0.00176,-0.00029]$ and actor-cluster interval
$[-0.00491,0.00141]$. The actor-specific deltas range from $-0.00674$ to
$+0.00176$; omitting actor 21 changes the pooled delta to $+0.00089$.
The same-actor statement delta is $+0.00004$, with pair-bootstrap interval
$[-0.00043,0.00052]$ and actor-cluster interval
$[-0.00017,0.00025]$. The new aligned, zero-audio, original actor-swap,
and identity-swap means agree with the registered replay within
$0.000001$ MAE. These post-confirmation controls are exploratory; neither their
pair-level interval nor a retrospective threshold changes the registered
decision, and four actor clusters do not support population inference.

To check whether the removal response corresponds to a useful fitted audio
signal on this raster endpoint, we separately fit the same AudioFiLM
architecture with a constant zero audio tensor for every example. We reused
the development and validation actor split, $64\times64$ target, seed 1701,
optimizer, batch size, 40-epoch maximum, patience, and validation-MAE
checkpoint selection. The comparator selected epoch 35; the original selected
epoch 18. Both checkpoints score the same 240 confirmation targets. The
constant-audio renderer obtains MAE 0.08652 versus 0.08990 for the original.
Its paired difference is $-0.00338$ MAE; the pair-bootstrap interval is
$[-0.00401,-0.00275]$ and the four-actor bootstrap interval is
$[-0.00512,-0.00164]$. For seed 1701, all four actor-specific mean differences are negative,
from $-0.00614$ to $-0.00143$. The original comparator was designed only after the confirmation actors had
been opened. Three further matched-seed comparisons used a fixed plan on that
same opened set.
Table~\ref{tab:ravdess-seeds} reports every seed; the constant-audio fit
has lower MAE in all four. The new training runs use seeds 1702--1704,
with all fit and selection settings held fixed except the seed and audio
input. They check the observed endpoint but cannot replace a prospectively
registered benefit test or imply perceptual video quality. One actor-specific
mean is positive in seed 1702 and two are positive in seed 1704; the latter
constant-audio fit selected the 40-epoch cap. Actor-level and run-to-run
uncertainty therefore remain.

We then repeated all four matched seed pairs with an 80-epoch maximum and the
same patience-six validation rule. Every constant-audio fit again has lower
MAE on the same targets, with differences ranging from $-0.00307$ to
$-0.00070$. For seed 1704, the constant-audio checkpoint now selects epoch
48, so this repeat's negative difference is not capped at epoch 40. None of
the new fits selects the 80-epoch cap. The new training histories differ from
the earlier histories before epoch 40, however; differences between runs
cannot be assigned solely to the longer budget. Table~\ref{tab:ravdess-seeds}
shows both sets without treating seeds or reruns as new test populations.

To separate continued fitting from differences between fresh runs, a further
frozen check retained the best validation checkpoint available by epoch 40
within each new 80-epoch trajectory, then scored that checkpoint and the
final selected checkpoint on the same 240 opened targets. Both arms selected
the same checkpoint at or before epoch 40 in seeds 1701--1703: aligned-audio
epochs 22, 15, and 19, and constant-audio epochs 34, 29, and 31. In seed
1704, aligned audio stayed at epoch 24, while constant audio selected epoch
38 by the old cap and epoch 48 after continuation. Its MAE fell from 0.08598
to 0.08538, an after-minus-before change of $-0.00060$. The
constant-minus-aligned difference moved from $-0.00298$ to $-0.00358$ in
that same trajectory. Actor-specific constant-audio changes range from
$-0.00310$ to $+0.00077$; this single retrospective fit does not establish
an actor-population effect. The check uses eight newly fitted trajectories
and the same opened confirmation clips. It does not change the registered
replacement result or convert the earlier independent 40-versus-80 fits
into a controlled budget comparison.

\begin{table}[ht]
\centering
\small
\caption{RAVDESS same-target sensitivity on 240 fixed targets.
Maximum epochs is the stopping cap; validation selection can stop earlier.
$\Delta$ is constant-audio-fit MAE minus aligned-audio-fit MAE, so negative
favors the constant-audio fit. The 80-epoch runs are fresh post-confirmation
fits, not exact continuations of the 40-epoch trajectories.}
\label{tab:ravdess-seeds}
\vspace{4pt}
\begin{tabularx}{0.91\linewidth}{@{}llccc@{}}
\toprule
\textbf{Seed} & \textbf{Max. epochs} & \textbf{Aligned MAE $\downarrow$} & \textbf{Constant MAE $\downarrow$} & \textbf{$\Delta$ MAE $\downarrow$} \\
\midrule
1701 & 40 & 0.08990 & 0.08652 & $-0.00338$ \\
1701 & 80 & 0.08803 & 0.08599 & $-0.00204$ \\
1702 & 40 & 0.08979 & 0.08749 & $-0.00230$ \\
1702 & 80 & 0.08874 & 0.08740 & $-0.00133$ \\
1703 & 40 & 0.08984 & 0.08719 & $-0.00265$ \\
1703 & 80 & 0.08841 & 0.08770 & $-0.00070$ \\
1704 & 40 & 0.08853 & 0.08637 & $-0.00216$ \\
1704 & 80 & 0.08873 & 0.08565 & $-0.00307$ \\
\bottomrule
\end{tabularx}
\end{table}

The earlier 1.0-second result is excluded because it does not measure the
registered 2.0-second endpoint; all reported RAVDESS results use the
2.0-second target and remain limited to held-out actor raster reconstruction.

\section{Development checks that stopped visible-speaker routes}
\label{app:stopped-routes}

These checks explain why the paper reports controlled raster and visual-proxy
outcomes instead of a positive visible-speaker result. None opened a sealed
confirmation set after its development or validation rule failed.

On AMI meeting recordings~\citep{amiCorpus}, a registered cross-session
word-timing positive control failed. A replacement test of participant headset
audio against close-up video found four eligible participants but only two
visual-timing positives in one opened session; a second session had only two
eligible participants. On HDTF clips~\citep{zhang2021flow}, a fixed audio-shift
timing test passed for two of four development clips and neither of two
validation clips. These outcomes stopped the respective source routes before
renderer fitting or confirmation; they are endpoint-eligibility findings, not
negative held-out renderer results.

On opened GRID development clips~\citep{cooke2006grid}, an AVA-style evaluator
favored aligned over shifted input for two clips and reversed for two; an
independent SyncNet checkpoint favored aligned input for one of four. A
source-local fixed-crop predictor then passed its directional check for four
of six validation talkers, below the preregistered five; its talker-bootstrap
interval $[-0.01798,0.06214]$ crossed zero. The reserved confirmation talkers
remained unopened. These development checks do not
supply a compatible external evaluator for a visible-speaker statement.

\end{document}